\documentclass[runningheads]{llncs}
\usepackage[T1]{fontenc}
\usepackage{graphicx}
\usepackage{multirow}%
\usepackage{amsmath,amssymb,amsfonts}%
\usepackage{mathrsfs}%
\usepackage[title]{appendix}%
\usepackage{xcolor}%
\usepackage{textcomp}%
\usepackage{manyfoot}%
\usepackage{booktabs}%
\usepackage{algorithm}%
\usepackage{algorithmicx}%
\usepackage{algpseudocode}%
\usepackage{listings}%
\usepackage{subcaption}
\usepackage{tikz}
\usepackage{placeins}
\usepackage{soul}
\usepackage{url}
\usepackage[numbers,sort&compress]{natbib}
\usepackage{hyperref}
\usepackage{caption}
\usepackage{tabularx}
\usepackage{pdfpages}
\usepackage[switch]{lineno}
\usepackage{xcolor, colortbl}
\usetikzlibrary{calc}
\usepackage{xspace}
\usepackage{array,booktabs}

\newcolumntype {+}{ >{\global\let\currentrowstyle\relax}}
\newcolumntype {^}{ >{\currentrowstyle }}

\newcommand{\ppms}{\textsc{PPMs}\xspace}
\newcommand{\ppm}{\textsc{PPM}\xspace}

\definecolor{colorp}{RGB}{0, 191, 192}      %
\definecolor{colorpa}{RGB}{128, 223, 223}   %
\definecolor{colorpb}{RGB}{191, 239, 239}   %
\definecolor{colorm}{RGB}{58,142,237}       %
\definecolor{colorma}{RGB}{156,198,246}     %
\definecolor{colorg}{RGB}{255,145,77}       %
\definecolor{colorga}{RGB}{255,199,166}     %
\definecolor{colorf}{RGB}{242,96,119}       %

\newcommand{\markp}[0]{\textcolor{colorp}{\rule{5pt}{5pt}}\xspace}      %
\newcommand{\markpa}[0]{\textcolor{colorpa}{\rule{5pt}{5pt}}\xspace}    %
\newcommand{\markpb}[0]{\textcolor{colorpb}{\rule{5pt}{5pt}}\xspace}    %
\newcommand{\markm}[0]{\textcolor{colorm}{\rule{5pt}{5pt}}\xspace}      %
\newcommand{\markma}[0]{\textcolor{colorma}{\rule{5pt}{5pt}}\xspace}    %
\newcommand{\markg}[0]{\textcolor{colorg}{\rule{5pt}{5pt}}\xspace}      %
\newcommand{\markga}[0]{\textcolor{colorga}{\rule{5pt}{5pt}}\xspace}    %
\newcommand{\markf}[0]{\textcolor{colorf}{\rule{5pt}{5pt}}\xspace}      %

\usepackage[most]{tcolorbox}
\newcommand{\summarybox}[2]{%
  \begin{tcolorbox}[
    colback=white,
    enhanced,
    frame code={
      \draw[gray!50!white, line width=0.5pt, rounded corners=4pt] (frame.south west) rectangle (frame.north east);
    },
    left=2pt, right=2pt, top=2pt, bottom=2pt, %
    overlay={
      \node[fill=#1, minimum size=10pt, inner sep=0pt] 
        at ([xshift=2pt,yshift=-2pt]frame.north west) {};
    },
    boxsep=5pt,
    rounded corners
  ]
  \sl #2
  \end{tcolorbox}%
}

\begin{document}
\title{This looks like what? Challenges and Future Research Directions for Part-Prototype Models}
\titlerunning{Challenges and Future Research Directions for Part-Prototype Models}
\author{Khawla Elhadri\inst{1} \and
Tomasz Michalski\inst{2,3} \and
Adam Wróbel\inst{2,3} \and
Jörg Schlötterer\inst{1} \and
Bartosz Zieliński\inst{3} \and
Christin Seifert\inst{1}}
\authorrunning{K. Elhadri et al.}
\institute{Marburg University, Germany \\
\email{\{khawla.elhadri,joerg.schloetterer,christin.seifert\}@uni-marburg.de}\\
\and
Jagiellonian University, Doctoral School of Exact and Natural Sciences, Poland
\email{\{tomasz.michalski,adam.wrobel\}@doctoral.uj.edu.pl}\\
\and
Jagiellonian University, Faculty of Mathematics and Computer Science, Poland\\
\email{bartosz.zielinski@uj.edu.pl}}
\maketitle              %
\begin{abstract}
The growing interest in eXplainable Artificial Intelligence (XAI) has stimulated research on models with built-in interpretability, among which part-prototype models are particularly prominent. Part-Prototype Models (PPMs) classify inputs by comparing them to learned prototypes and provide human-understandable explanations of the form “this looks like that.” Despite this intrinsic interpretability, PPMs have not yet emerged as a competitive alternative to post-hoc explanation methods.
This survey reviews work published between 2019 and 2025 and derives a taxonomy of the challenges faced by current PPMs. The analysis reveals a diverse set of open problems. The main issue concerns the quality and number of learned prototypes. Further challenges include limited generalization across tasks and contexts, as well as methodological shortcomings such as non-standardized evaluation.
Five broad research directions are identified: improving predictive performance, developing theoretically grounded architectures, establishing frameworks for human–AI collaboration, aligning models with human concepts, and defining robust metrics and benchmarks for evaluation. The survey aims to stimulate further research and promote intrinsically interpretable models for practical applications.
A curated list of the surveyed papers is available at \url{https://github.com/aix-group/ppm-survey}.

\keywords{Part-prototype models \and ante-hoc interpretability \and survey.}
\end{abstract}
\begin{figure*}[tbh]
    \centering
\includegraphics[width=\textwidth]{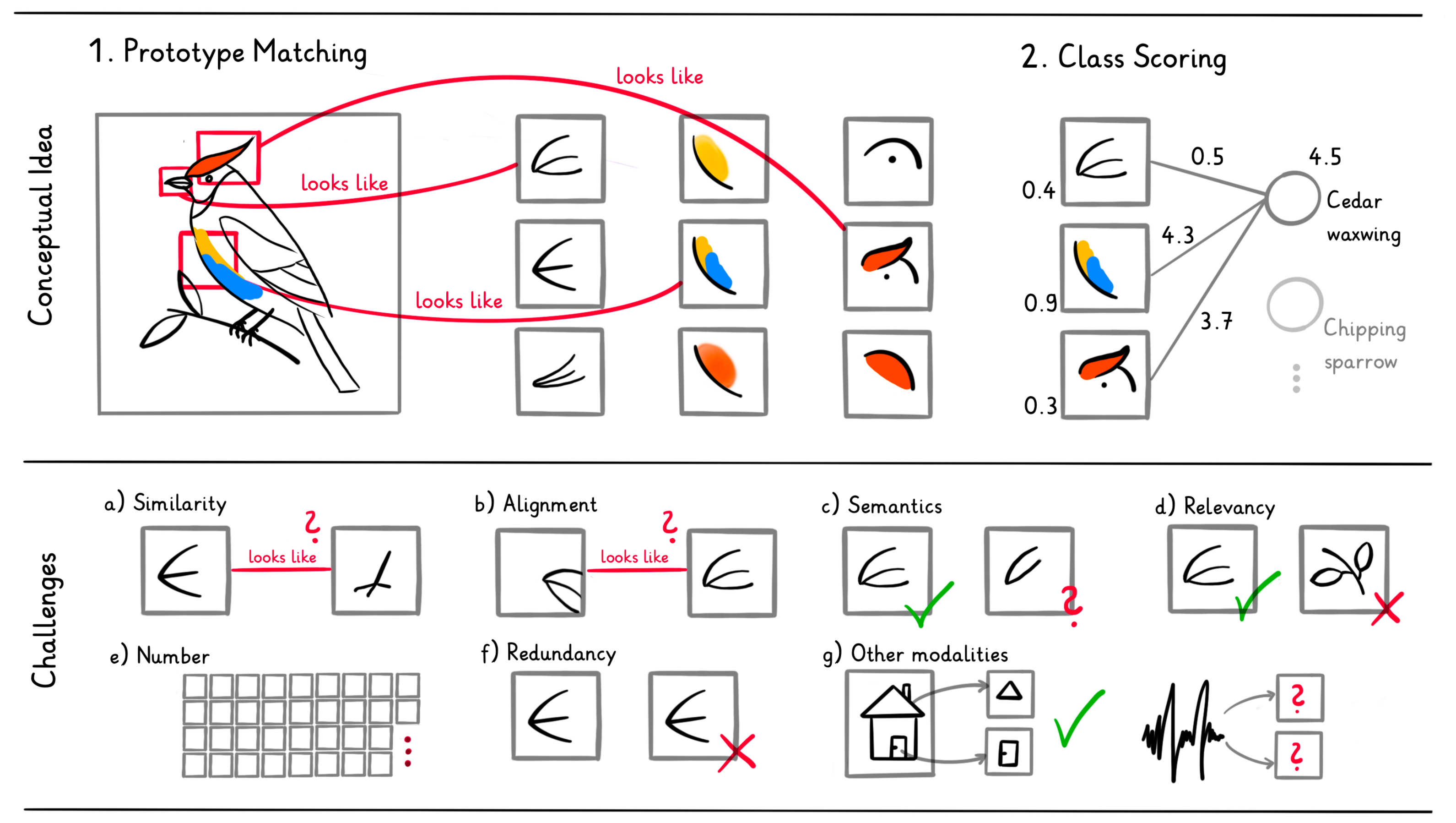}
\caption{Overview of the reasoning process of part-prototype models (\ppms) (top) and selected challenges (bottom).
\ppms match learned prototypes (e.g., birds’ beaks, breasts, and heads) to patches in the input image by computing patch–prototype similarities. The decision layer combines these similarities with learned class weights through a simple scoring scheme.
Key challenges remain: the model may not explain why it matches a patch to a specific prototype (a); matches may misalign spatially (b); prototypes may lack human meaning (c) or represent task-irrelevant concepts (d). The model may also learn too many (e) or redundant prototypes (f), and it remains unclear how to define and interpret prototypes for non-image modalities such as sensor data or text (g).}
    \label{fig:overview}
\end{figure*}

\begin{figure*}[p]
    \centering
 \includegraphics[width=\textwidth]{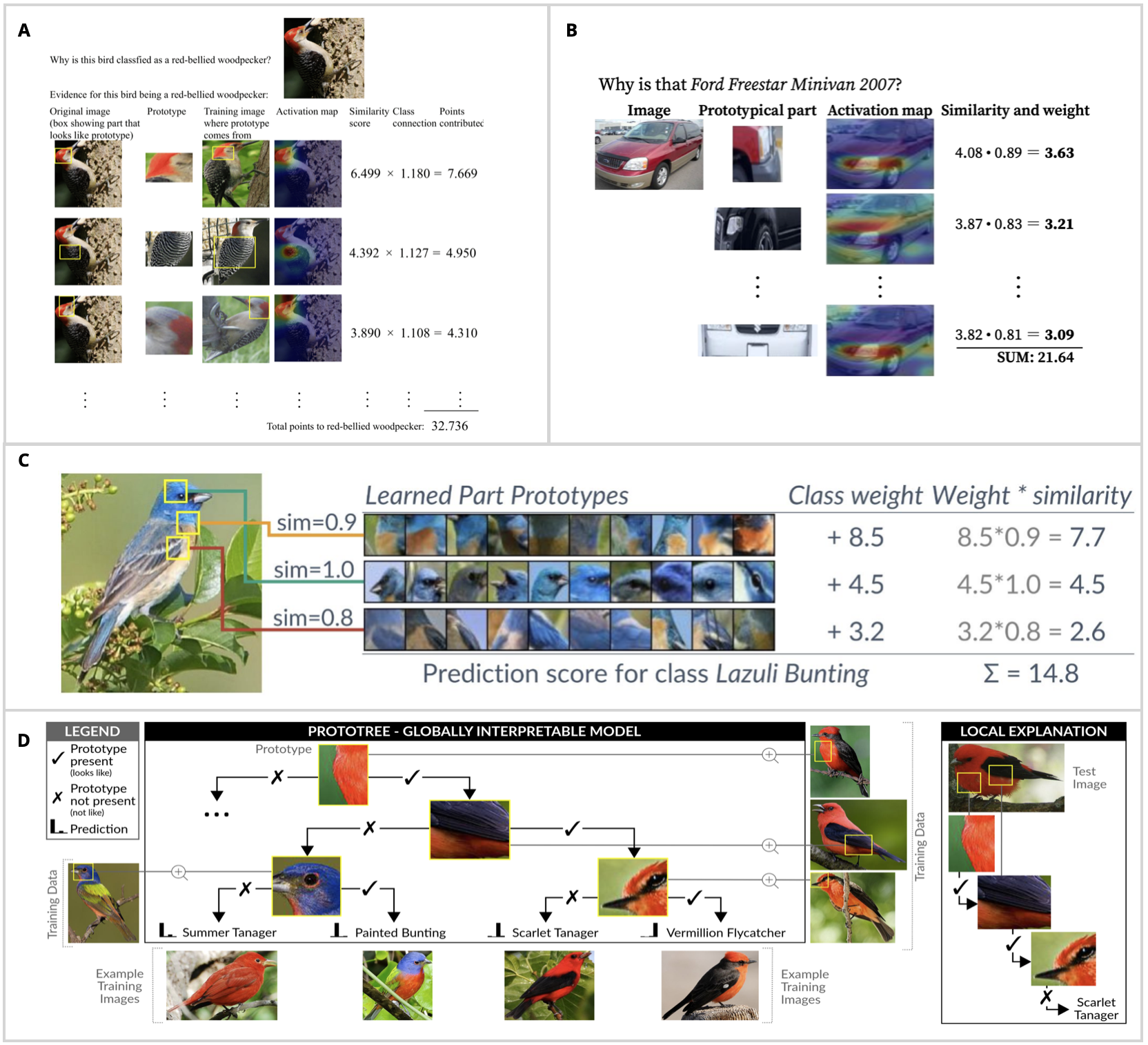}
\caption{Examples of reasoning in part-prototype models. 
(\textbf{A}) \textsc{ProtoPNet}~\citep{Chen_2019_ThisLooksThat} explains bird classification by matching patches of the test image to learned class-specific prototypes, yielding case-based explanations of the form “this part of the image looks like that prototypical part.” The figure shows, for each class, the prototype patches from training images, the corresponding activation maps on the input highlighting the best-matching regions, the resulting similarity scores, and the class weights used to compute each prototype’s contribution to the final prediction.
(\textbf{B}) \textsc{ProtoPool}~\citep{Rymarczyk_2022_InterpretableImageClassification} follows the same weighted-sum principle but draws prototypes from a shared pool across classes, explaining a vehicle prediction through the most activated prototypical parts and their class-specific weights. 
(\textbf{C}) \textsc{PiP-Net}~\citep{Nauta_2023_PIPNetPatchBasedIntuitive} also aggregates weighted prototype similarities but can abstain when no confident prototype match is found. 
(\textbf{D}) \textsc{ProtoTree}~\citep{Nauta_2021_NeuralPrototypeTrees} uses a different decision structure: a globally interpretable tree in which internal nodes test for the presence or absence of learned prototypical parts, producing predictions through an explicit and human-readable decision path.}
\label{fig:ppm-reasoning}
 \end{figure*}

\begin{figure*}[ht]
    \centering
 \includegraphics[width=\textwidth]{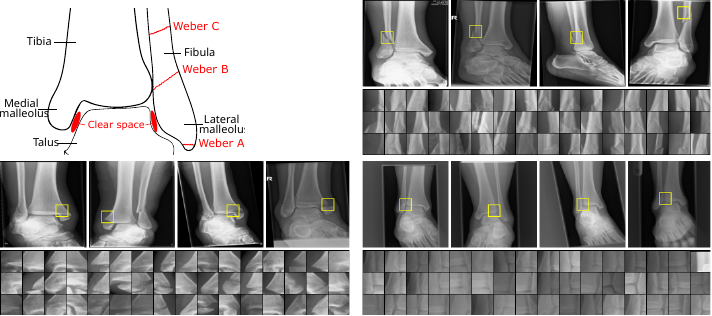}
     \caption{\textsc{PiP-Net} for fracture detection~\citep{Nauta_2024_InterpretingCorrectingMedical} learns clinically meaningful prototypes aligned with Weber fracture types: a Weber B prototype, (top right), a prototype for the distal medial malleolus (bottom left), and a prototype relevant for no-fracture cases located in a fracture-prone region (bottom right).}
    \label{fig:medical_use_pipnet}
\end{figure*}

Machine learning systems are increasingly adopted in high-stake domains, from autonomous vehicles (e.g.,~\cite{Liao_2024_CDSTraj}), to finance (e.g.,~\cite{Zhou_2024_Astrategicanalysis}), and healthcare (e.g.,~\cite{Eisemann_2025_Nationwiderealworldimplementation}). 
Operating in areas where trustworthiness is expected~\citep{Li_2023_Trustworthyai}, these systems are required to provide explanations for their decisions. This need for transparency has motivated the rise of eXplainable Artificial Intelligence (XAI). Initially, the focus of XAI has been on building post-hoc methods that explain the reasoning process of already built (black box) models. However, post-hoc methods only approximate the model's predictions without ever reaching perfect faithfulness~\citep{Rudin_2019_Stopexplainingblack}. Thus, a new category of models has emerged that are interpretable by design and transparent about their decision-making process. A prominent class of such ante-hoc models are part-prototype models (\ppms).
To make their predictions, \ppms compare the input data to prototypical parts learned during training. 
Prototypical parts correspond to visual concepts, such as pointy ears, a hooked beak in natural images of animals, or vascular calcification in breast cancer x-rays.
After a \ppm identifies image regions corresponding to part-prototypes, it makes its prediction by combining the similarities between these prototypes and the matched patches. Figure~\ref{fig:overview} (top) shows this process schematically, and Figure~\ref{fig:ppm-reasoning} presents concrete examples of \ppms’ reasoning. Figure~\ref{fig:medical_use_pipnet} illustrates PiP-Net in a medical application.

Despite their intrinsic interpretability, \ppms are not yet widely adopted, and often secondary to black box models~\citep{Rudin_2019_Stopexplainingblack}. 
To better understand this gap, we surveyed recent work (including methods and analysis papers) on \ppms and analyzed open problems and challenges mentioned by the authors. We systematically collected papers from 17 premier venues, published between 2019 and 2025. 
An in-depth analysis of these 53 papers resulted in a taxonomy of challenges. More specifically, we found that current \ppms still face four challenges: 
1)  Learned prototypes
\footnote{A prototype represents a whole object, whereas prototypical parts (part-prototypes) represent object parts. For brevity, we use the term ``prototypes'' throughout.}
have shortcomings that hinder the interpretability of \ppms, which is a major barrier to their adoption as an alternative to black box models (\textbf{Prototypes}, Section~\ref{ssec:chall:proto}); 
2) \ppms lack a theoretical foundation, face training and inference challenges, and lack standardized evaluation and benchmarks (\textbf Methodology, Section~\ref{ssec:chall:method}); 
3) \ppms are underexplored beyond fine-grained image recognition and in settings with fewer assumptions (\textbf{Generalization}, Section~\ref{ssec:chall:gen});
4) \ppms have limitations that prevent them from being used in practice, even if the other challenges were solved (\textbf{Safety \& Use in Practice}, Section~\ref{ssec:chall:safety}).

Based on a detailed analysis of challenges, we identified five main research directions and provide more detailed ideas on how to make a significant contribution to the successful application of \ppms in practice:
1) Improve predictive performance of \ppms to make them a valuable interpretable alternative to high-performing black-box models (\textbf{Performance}, Section~\ref{ssec:directions:performance});
2) Design novel \ppm-based architectures for different modalities taking theoretical understanding of concepts for those modalities into account (\textbf{Novel Architectures}, Section~\ref{ssec:directions:architectures}); 
3) Design frameworks to allow interactive exploration and adaptation of \ppms (\textbf{Human-AI Collaboration}, Section~\ref{ssec:directions:interactive});
4) Align semantics of prototypes and \ppms reasoning with human expectations without impairing model performance (\textbf{Model Alignment}, Section~\ref{ssec:directions:alignment});
5) Establish comprehensive benchmarks including synthetic and real-world datasets and metrics for evaluating explanation quality, both generally and for specific application domains (\textbf{Metrics and Benchmarks}, Section~\ref{ssec:directions:benchmarks}).

\paragraph{Relation to other Surveys.} 
In this survey, we focus on \ppms as a growing category of ante-hoc models that are gaining interest in modalities beyond computer vision due to their intuitive, example-based explanations.
To our knowledge, no other survey focuses specifically on \ppms. Existing XAI surveys mention \ppms in the context of interpretability in ante-hoc methods~\cite{2025_Marino_AnteHocMethods}, the interpretability–accuracy trade-off~\citep{Ibrahim2022_acm-csur_Explainable-CNNs}, evaluation~\citep{Nauta2023_csur_evaluating-xai-survey,Nauta2023_wcxai_co-12-for-prototype-models}, or application domains~\citep{Alpherts2024_facct_perceptive-visual-urban-analytics,Patricio2024_acm-csur_XAI-medical-image-classification}. However, their assessment of \ppms is  limited to a handful of methods. Furthermore, they neither systematically analyze open challenges in \ppms nor consider non-vision modalities. This survey addresses this gap by providing a detailed, fine-grained and higher coverage analysis of \ppms, addressing their open challenges, and outlining directions for future research.

\paragraph{Survey Organization.} 
Figure~\ref{fig:overview} and Section~\ref{sec:background} introduce part-prototype models (\ppms). Section~\ref{sec:method} describes our survey method. Figure~\ref{fig:challenges-taxonomy} and Section~\ref{sec:challenges} present our taxonomy of open challenges. Section~\ref{sec:directions}, with Figure~\ref{fig:directions}, outlines future research directions, and Section~\ref{sec:conclusion} concludes.

\section{Background on Part-Prototype Models}
\label{sec:background}
We introduce part-prototype models (\ppms) using the seminal ProtoPNet~\citep{Chen_2019_ThisLooksThat}. ProtoPNet follows the principle of ''this looks like that``, comparing an input image to learned prototypes and basing its prediction solely on these comparisons. Figure~\ref{fig:overview} (top) illustrates this idea, and Figure~\ref{fig:ppm-reasoning}A shows a concrete example.

\subsection{ProtoPNet}
ProtoPNet consists of a convolutional neural network (CNN) to encode the input, a prototype layer to match the input with the prototypes, and a fully connected layer to compute the final decision.

The CNN is trained to map the input image onto a latent space of dimensionality $W\times H\times D$. The prototype layer acts as a bottleneck and consists of $K$ prototype representations, each of dimensionality $1\times 1\times D$. 
The representations of all $K$ prototypes are compared to the representations at each of the $W\times H$ locations of the CNN's latent grid.
These latent grid locations can be mapped back to pixel regions (patches) in the input image. 
Thus, the prototype layer outputs similarity scores that reflect the similarity of all input patches to all prototypes.
The highest similarity score per prototype (corresponding to a particular input patch) is passed to the final layer, which computes the classification output.
The weights of the final layer are constrained to be positive numbers, resulting in a decision that is a positive linear combination of prototype similarities, acting as a scoring sheet.

ProtoPNet is trained iteratively in three steps: i) stochastic gradient descent (SGD) of the CNN and the prototype layer with the final layer kept frozen), ii) projection of prototypes to training image patches, and iii) convex optimization of the final layer with the CNN and prototype layer frozen. The training loss minimizes the classification loss and encourages representations of latent patches to be close to prototype of their class, and far from prototypes of other classes.
The projection in step ii) sets the prototype representation to the representation of the training patch closest in latent space.

ProtoPNet is considered inherently interpretable, because (i) the decision layer is a simple linear model that is easy to analyze, and (ii) the decision is based solely on the prototypes, which in turn (iii) reflect meaningful and representative parts of the data (see Figure~\ref{fig:overview}).

\subsection{Extensions}
Multiple extensions of ProtoPNet have been introduced, e.g., to further improve the interpretability of prototypes by ensuring their disentanglement~\citep{Wang_2021_InterpretableImageRecognition}, making them context aware~\citep{Donnelly_2022_DeformableProtoPNetInterpretable} or grouping them in the latent space~\citep{Ma_2023_ThisLooksThose}. 
Also with the goal to ease interpretability, multiple authors focus on limiting the number of prototypes through adopting class-agnostic prototypes~\cite{Rymarczyk_2021_ProtoPSharePrototypicalParts,Nauta_2021_NeuralPrototypeTrees,Rymarczyk_2022_InterpretableImageClassification}. 
Further extensions include the integration of prototypes into transformer architectures~\citep{Xue_2024_ProtoPFormerConcentratingPrototypical,Ma_2024_Interpretableimage} and the application to other modalities beyond vision (e.g.,~\cite{Wang_2023_PROMINETPrototypebasedMultiView}).

\section{Survey Method}
\label{sec:method}

\begin{figure}[t!]
\centering
\small

\begin{minipage}{0.3\linewidth}
\centering
\begin{tabular}{lc}
\toprule
Modality & Count \\
\midrule
Images & 42 \\
Text   & 2  \\
Seq.   & 4  \\
Graph  & 1  \\
Sound  & 1  \\
Video  & 1  \\
Vision-Langauge & 1 \\
Multimodal & 1 \\
\midrule
Total  & 53 \\
\bottomrule
\end{tabular}\\
\end{minipage}
\hfill
\begin{minipage}{0.65\linewidth}
\centering
\includegraphics[width=\linewidth,trim={1cm 0.9cm 0 0},clip]{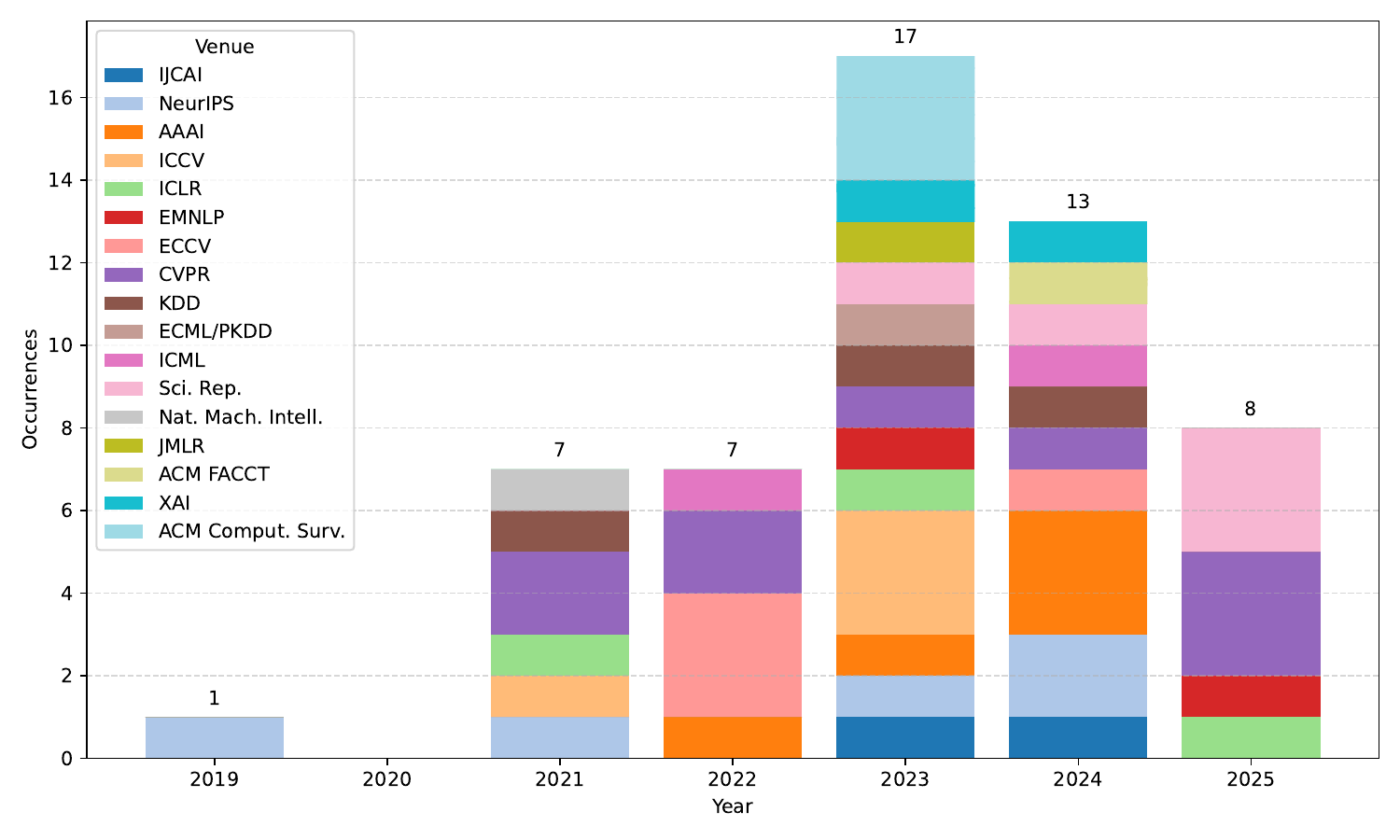}
\end{minipage}

\caption{Corpus overview. Left: Number of papers per modality (Seq. - Sequences). 
Right: Number of papers per venue and year. Survey papers are marked with a star pattern. 
Note: The seed paper on ProtoPNet~\protect\citep{Chen_2019_ThisLooksThat} was the only paper published in 2019, and first subsequent work appeared in 2021 in our corpus.}
\label{fig:chart}
\end{figure}

We conducted a structured search with ProtoPNet~\citep{Chen_2019_ThisLooksThat} as a seed paper to build our corpus of papers on part-prototype models (\ppms). 
We chose ProtoPNet because it is the first paper to introduce a neural architecture based on prototypical parts. 
We filtered the corpus by the following inclusion and exclusion criteria:
\begin{itemize}
    \item We only considered premier venues\footnote{CORE ranking at least A or topically relevant, but yet unranked. Upon an initial query, publications in the following venues matched our other inclusion criteria, constituting the final list: IJCAI, NeurIPS, AAAI, ICCV, ICLR, EMNLP, ECCV, CVPR, KDD, ECML/PKDD, ICML, Sci. Rep., Nat. Mach. Intell., JMLR, ACM FAcct, XAI, ACM Comput. Surv.} and excluded workshop papers, posters, and shared tasks.
    \item We included papers that present \ppms, namely novel methods, methods that improve on or apply ProtoPNet, applications of existing \ppms, surveys, evaluation with humans, and/or evaluation frameworks.
\end{itemize}

We queried the Semantic Scholar API\footnote{\tiny\url{https://api.semanticscholar.org/graph/v1/paper/cc145f046788029322835979a14459652da7247e/citations?fields=intents,url,title,abstract,venue,year,referenceCount,citationCount,influentialCitationCount,fieldsOfStudy,publicationDate&limit=1000} and a second call with offset=1000 on 28.01.2026} for all papers that cite ProtoPNet and were published from 2019 to 2025, resulting in 1395 papers.
We filtered this set by the selected venues and exclusion criteria. We conducted a web search for papers with missing venue information (106 papers) and for peer-reviewed versions of papers with arXiv as venue (291) and ``proto'' in their (lowercased) title (54).
We then manually reviewed each paper for our inclusion criteria.

Our final corpus consists of 53 papers, the majority of which are on image processing (see Figure \ref{fig:chart}), including 45 methods papers, 3 analysis papers, and 5 surveys.
As shown in Figure~\ref{fig:chart}, research on \ppms is published in diverse venues and has increased steadily until 2023, then gradually decreased in 2024 and 2025. This decrease could be due to several reasons: We included surveys, four of which were published in 2023. We relied on publications that cited the seminal paper by~\cite{Chen_2019_ThisLooksThat}, but there may be papers presenting, applying or evaluating \ppms that do not cite this paper. In addition, the venue information provided by Semantic Scholar's API is not always accurate and may have affected our initial corpus. 
Another reason may be that the complexity of the open challenges and the lack of clear future directions have stagnated research in \ppms.

\section{Challenges of Part-Prototype Models}
\label{sec:challenges}

\begin{figure*}[tbp]
    \centering
 \includegraphics[width=\textwidth]{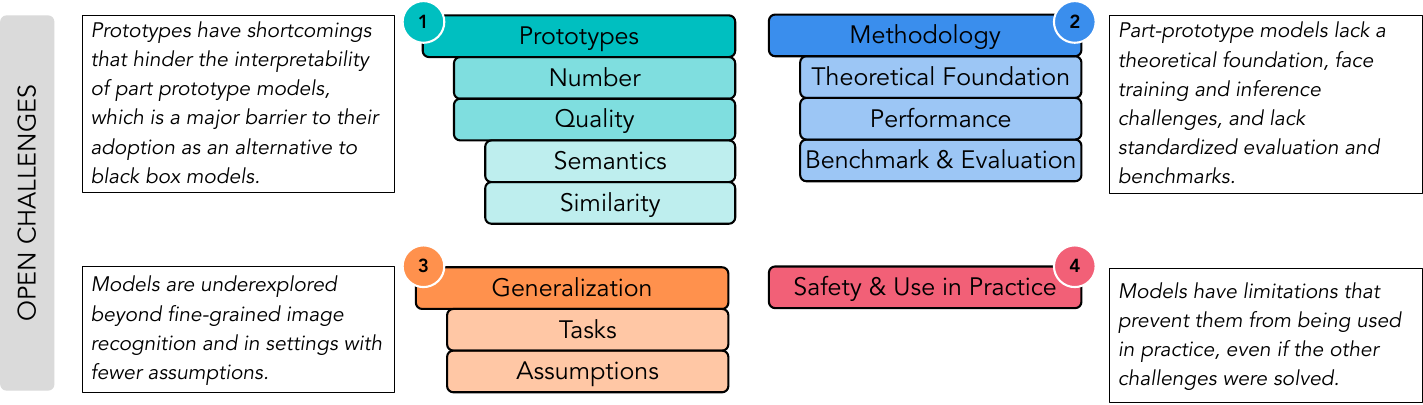}
    \caption{Taxonomy of challenges for part-prototype models.}
    \label{fig:challenges-taxonomy}
\end{figure*}

In this section, we present our taxonomy of open challenges for part-prototype models (\ppms) with four main categories (see Figure~\ref{fig:challenges-taxonomy}).
First, we outline the challenges related to the number and quality of prototypes (category \texorpdfstring{\markp 
\textbf{Prototypes}}{\textbf{Prototypes}} in Section~\ref{ssec:chall:proto}). Second, we describe the challenges related to the theoretical foundation of \ppms, their performance, and the lack of standardized evaluation (category \texorpdfstring{\markm 
\textbf{Methodology}}{\textbf{Methodology}} in Section~\ref{ssec:chall:method}). Third, we examine the limitations of \ppms with respect to the machine learning tasks they have been applied to and the assumptions determining their architecture (category \texorpdfstring{\markg 
\textbf{Generalization}}{\textbf{Generalization}} in Section~\ref{ssec:chall:gen}). 
Fourth, we point out concerns that prevent \ppms from being used in practice (category \texorpdfstring{\markf 
\textbf{Safety and Use in Practice}}{\textbf{Safety and Use in Practice}} in Section~\ref{ssec:chall:safety}).
We outline ideas from the papers in our surveyed corpus to address some of these open challenges in Tables \ref{tab:ideas:prototypes} and \ref{tab:ideas:method-general}.

\subsection{\texorpdfstring{\markp Prototypes}{Prototypes}}
\label{ssec:chall:proto}
The interpretability of \ppms is governed by the number and the quality of learned prototypes, namely their quantity and the extent to which their semantics are human-understandable.

\subsubsection{\texorpdfstring{\markpa Number of Prototypes}{Number of Prototypes}.}
\label{sssec:chall:proto:number}
In the \ppm architecture, the number of prototypes determines the size of the bottleneck layer.
\footnote{The argument also holds for more complex building blocks than single-layers.}
Finding the optimal number of prototypes is therefore similar to finding an optimal neural architecture~\citep{Elsken2019_survey-neural-arch-search}.
Early prototype models fixed the number of prototypes as a multiple of the number of classes (e.g., \cite{Chen_2019_ThisLooksThat,Li_2024_ImprovingPrototypicalVisual}). This results in many redundant prototypes~\citep{Davoodi_2023_interpretabilitypartprototypebased} (see Figure \ref{fig:overview}f), because some object parts may discriminate between sets of classes rather than single classes. 
Therefore, extensions aim to reduce duplicate prototypes by pooling~\citep{Rymarczyk_2022_InterpretableImageClassification}, pruning~\citep{Rymarczyk_2021_ProtoPSharePrototypicalParts}, hierarchical ordering~\citep{Nauta_2021_NeuralPrototypeTrees}, or regularization on the number of prototypes~\citep{Nauta_2023_PIPNetPatchBasedIntuitive}.

While these approaches address the problem of prototype redundancy, they do not identify the optimal number of prototypes.
A very small number of prototypes is easier for humans to interpret, but a larger number allows the model to learn more semantically meaningful prototypes~\citep{Davoodi_2023_interpretabilitypartprototypebased} (see Figure~\ref{fig:overview}e).

\subsubsection{\texorpdfstring{\markpa Quality of Prototypes}{Quality of Prototypes}.}
\label{sssec:chall:proto:quality}
The quality of prototypes has two  aspects: \textit{Semantics}, i.e., whether a prototype is relevant to the task and understandable by humans, and \textit{Similarity}, i.e., whether the mapping from the image part to the prototype makes sense to humans.

\begin{figure*}[tbp]
    \centering
 \includegraphics[width=\textwidth]{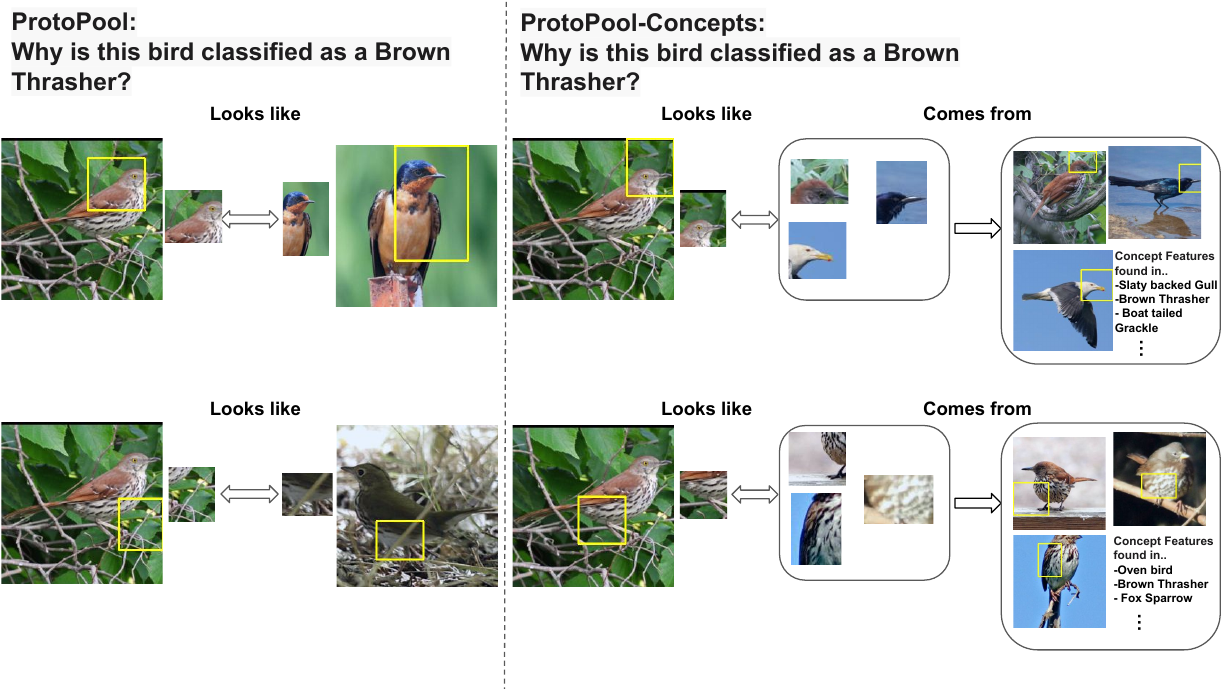}
 \caption{Example interpretability gap in \ppms~\citep{Ma_2023_ThisLooksThose}. For a Brown Thrasher, \textsc{ProtoPool} (left) shows ambiguous single prototypes, while \textsc{ProtoPool-Concepts} (right) provides multiple examples that clarify visual patterns but still lack explicit semantic labels. This highlights a key challenge: clearer visual evidence does not resolve the human-dependent meaning of prototypes.}
    \label{fig:semantic_alignment}
\end{figure*}

\paragraph{\texorpdfstring{\markpb Semantics}{Semantics}.}
\label{ssssec:chall:proto:quality:sem}
The interpretability of \ppms relies largely on the quality of the prototypes, currently hindered by the following semantic challenges:

\paragraph{Prototype Interpretability.}
\ppms classify an input image by comparing patches of the image to learned prototypes (see Prototype Matching in Figure \ref{fig:overview}). This matching does not specify what the model looks at (shape, color, texture) to assess similarity~\citep{Nauta_2021_NeuralPrototypeTrees}.
One approach to this problem is representing a prototype as multiple image parts (patches) instead of one~\citep{Ma_2023_ThisLooksThose} as shown in Figure~\ref{fig:semantic_alignment}.
While insightful, this approach still relies on the user's own cognitive skills to infer which concept is being compared. Similarly, PIP-Net~\citep{Nauta_2023_PIPNetPatchBasedIntuitive} learns prototypes that represent semantically meaningful concepts and align with human intuition, but explicit semantics remain missing. In recent work, LucidPPN~\citep{Pach_2024_LucidPPNUnambiguousPrototypical} introduces color as semantic explanation to the model's input image-prototypes matching. While some semantic clarity is introduced, other relevant features such as shape and texture remain overlooked. 

\paragraph{Prototype Information.} 
Early \ppms tend to learn prototypes that focus on background information, due to the inadvertently learned dataset biases~\citep{Chen_2019_ThisLooksThat,Nauta_2021_NeuralPrototypeTrees}.
Later work used humans-in-the-loop to adjust the learned prototype post-hoc, by removing confounded prototypes upon human inspection~\citep{Bontempelli_2023_ConceptlevelDebuggingPartPrototype}. 
Similarly, Li et al.~\cite{Li_2024_ImprovingPrototypicalVisual} re-weight, re-select and re-train prototypes using a reward model that is trained on human preference feedback. However, both methods require the model to learn unambiguous concepts because ambiguous prototypes (i.e., prototypes that represent multiple concepts at once) make human evaluation difficult. 
In medical applications, \ppms struggle to accurately capture the region of interest\footnote{The most prediction-relevant concepts in an image.}~\citep{Pathak_2024_PrototypebasedInterpretableBreast}. This manifests in many irrelevant prototypes (see Figure~\ref{fig:overview}d), few pure\footnote{Nauta et al.~\cite{Nauta_2023_PIPNetPatchBasedIntuitive} defines purity as ``the fraction of image patches of a prototype that have overlap with the same ground-truth object part''.} and unique prototypes, and a spatial misalignment between the input image and visualization of prototypes (see Figure~\ref{fig:overview}b).

\begin{figure}[thbp]
    \centering
    \begin{tikzpicture}
        \def\imageheight{4.5cm}
        \def\imagewidth{0.48\textwidth}

        \node[anchor=south west, inner sep=0] (imageA) at (0,0)
            {\includegraphics[width=\imagewidth,height=\imageheight,keepaspectratio]{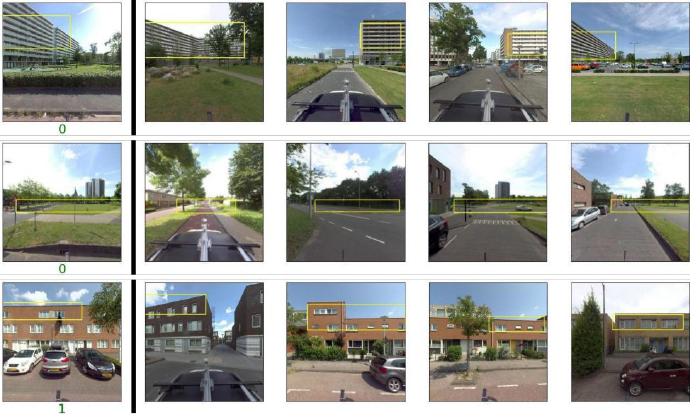}};

        \node[anchor=south west, inner sep=0] (imageB) at ([xshift=0.04\textwidth]imageA.south east)
            {\includegraphics[width=\imagewidth,height=\imageheight,keepaspectratio]{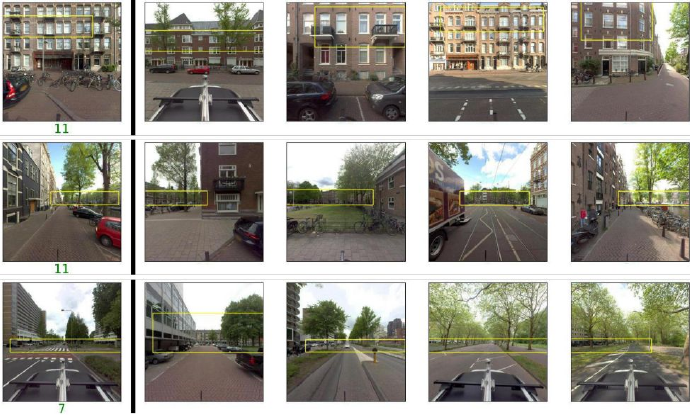}};

        \draw[line width=0.4pt]
            ($(imageA.south east)+(0.02\textwidth,0)$) --
            ($(imageA.north east)+(0.02\textwidth,0)$);
        \node at ($(imageA.south)+(0,-0.4)$) {(a) Lower price neighborhoods};
        \node at ($(imageB.south)+(0,-0.4)$) {(b) Higher price neighborhoods};
    \end{tikzpicture}
    \caption{
    Examples of \textsc{ProtoPNet} prototypes for (a) lower and (b) higher housing price neighborhoods~\citep{Alpherts2024_facct_perceptive-visual-urban-analytics}. Large patches cover multiple elements, obscuring which features drive the decision.}
    \label{fig:visualization_protopnet}
\end{figure}

\paragraph{Prototype Visualization.} 
A prototype is a vector in latent space. It is visualized by the rectangular patches from the training set, whose representations are close to the prototype. However, using rectangular patches to visualize prototypes does not always favor interpretability, 
as it encompasses multiple visual concepts, which can cause confusion in understanding exactly what the model is highlighting and introduces the risk of a confirmation bias~\citep{Alpherts2024_facct_perceptive-visual-urban-analytics} (see Figure~\ref{fig:visualization_protopnet}). While single-modality prototypes can be visualized, this is not possible in MMPNet~\cite{Song_2025_Multimodalprototypicalnetwork}.
The prototypes represent patterns across several modalities. As each prototype encodes a temporal multimodal information, it cannot be reconstructed into human interpretable representations in the input space.

\begin{figure*}
    \centering
 \includegraphics[width=\textwidth]{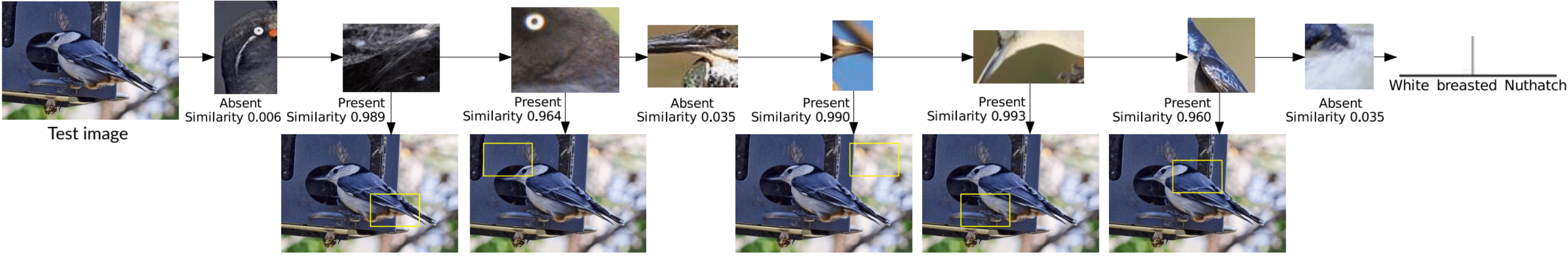}
\caption{Example from Kim et al.~\cite{Kim_2022_HIVEEvaluatingHuman} illustrating the similarity perception gap in \textsc{ProtoTree}~\citep{Nauta_2021_NeuralPrototypeTrees}. The model classifies a test bird image through a sequence of binary decisions based on prototype comparisons, assigning high similarity scores (ranging from 0.960 to 0.993) to prototypes that may appear semantically dissimilar to human observers. This demonstrates how the model's latent space similarity judgments do not align with human intuitions about visual similarity. This fundamental challenge of bridging the gap between computational and human similarity perception remains unresolved in current prototype-based interpretability methods.}
    \label{fig:similarity_hive}
\end{figure*}

\paragraph{\texorpdfstring{\markpb Similarity}{Similarity}}
\label{ssssec:chall:proto:quality:sim}
\ppms infer similarity between part of the input and a prototype by comparing vector representations in latent space and calculating a similarity score. However, such a score lacks explicit and clear semantics for the inferred similarity~\citep{Hong_2023_ProtoryNetInterpretableText}. %
In particular, when visualized in the input space, prototypes may seem to activate on input parts that are semantically dissimilar for humans~\citep{Donnelly_2022_DeformableProtoPNetInterpretable,Hong_2023_ProtoryNetInterpretableText}. This gap in similarity perception is quantified by Kim et al.~\cite{Kim_2022_HIVEEvaluatingHuman}, revealing a clear misalignment in judgment between humans and \ppms (see Figure~\ref{fig:similarity_hive}).

\summarybox{colorp}{{\bf Summary of Challenge:} 
 It remains unknown how to determine the optimal number of prototypes, i.e., how to balance interpretability (fewer prototypes) and predictive performance (more prototypes). High-quality prototypes should be diverse, represent semantically meaningful concepts that humans understand, and activate on the part of the input that humans perceive as semantically similar.}

\subsection{\texorpdfstring{{\markm Methodology}}{Methodology}}

\label{ssec:chall:method}
From training and inference to evaluation and theoretical analysis, the development of \ppms poses multiple methodological challenges for the community.

\subsubsection{\texorpdfstring{\markma Theoretical Foundation}{Theoretical Foundation}.}
\label{sssec:chall:method:tf}
Recent work on \ppms contains less theoretical analysis than earlier papers, such as~\citep{Chen_2019_ThisLooksThat}. Even when theory does appear, it is limited to properties of prototypical part representations, after the projection phase.
Hong~\cite{Hong_2023_ProtoryNetInterpretableText} calls for a mathematical formalization and enforcement of well-established requirements for linguistic prototypes, but we consider it to be a generally relevant desideratum for all kinds of prototypes. In the example of a linguistic prototype, according to the conditions proposed by Panther \& Köpcke~\cite{panther2008prototype}, the prototype must be an affirmative declarative sentence, where the subject is in the nominative case, the verb in the active voice and in the indicative mood.

\subsubsection{\texorpdfstring{\markma Performance}{Performance}.}
\label{sssec:chall:method:perf}

Despite their advantages in terms of interpretability, \ppms suffer from practical limitations that affect training and inference.
Training consists of several steps in which different parts of these models are trained~\citep{Zhang_2022_ProtGNNSelfExplainingGraph}. It often uses a large number of hyperparameters~\citep{Ruis_2021_IndependentPrototypePropagation,Ma_2024_Interpretableimage}, which require careful tuning~\citep{Rymarczyk_2022_InterpretableImageClassification}.

Therefore, training may take longer~\citep{Alpherts2024_facct_perceptive-visual-urban-analytics} and be less stable than in the case of black box models. In addition, inadequate regularization~\citep{Bontempelli_2023_ConceptlevelDebuggingPartPrototype} can lead to overfitting and poor generalization~\citep{Carmichael_2024_ThisProbablyLooks}.
During inference, these models often process input more slowly than black box models~\citep{Fauvel_2023_LightweightEfficientExplainablebyDesign}. Moreover, their performance degrades in the presence of noise or image transformations~\citep{Patricio2024_acm-csur_XAI-medical-image-classification,Nauta2023_wcxai_co-12-for-prototype-models}.
Finally, due to the randomness in training, there are significant inconsistencies in explanations generated across different runs (i.e., different prototypes are found across different runs)~\citep{Nauta2023_wcxai_co-12-for-prototype-models}.

\subsubsection{\texorpdfstring{\markma Benchmark and Evaluation}{Benchmark and Evaluation}.}
\label{sssec:chall:method:bench}
\begin{figure}[tbhp]
 \centering
\includegraphics[width=\textwidth]{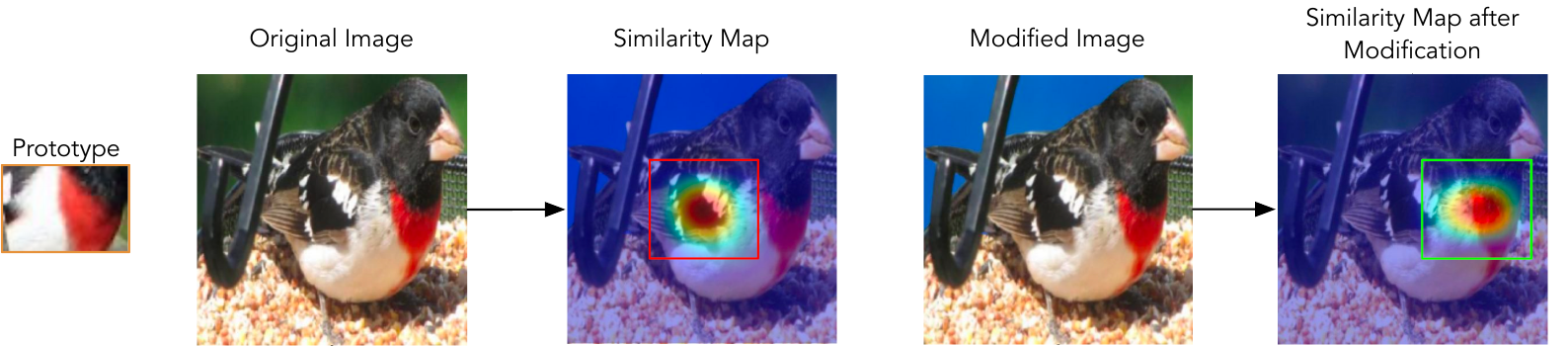}
        \caption{The Spatial Misalignment benchmark~\citep{Sacha_2024_Interpretabilitybenchmarkevaluating} tests whether prototype-based explanations align with the image regions that drive model decisions by measuring changes in prototype activations under background perturbations. It exposes spatial misalignment when explanations highlight one region but the model relies on another. However, as a proxy measure, it may not generalize to more complex settings such as comparisons across multiple models or ensembles.}
    \label{fig:benchmark_1}
\end{figure}

\ppms are evaluated in two aspects. The first is performance, as with standard models, and the second is explainability.
In terms of performance, \ppms are typically compared on well-formed (balanced, IID) benchmarks using accuracy as a performance measure, which lacks the realistic challenges, such as out-of-distribution data, for which overconfident estimates are often obtained~\citep{Nauta2023_wcxai_co-12-for-prototype-models}.
The second aspect, explainability, is assessed both quantitatively and qualitatively. Quantitative analysis assesses various aspects of explanations through automated proxy measures, such as their consistency across images and robustness~\citep{Huang_2023_EvaluationImprovementInterpretability}, or spatial misalignment~\citep{Sacha_2024_Interpretabilitybenchmarkevaluating}. An example for spatial misalignment is shown in Figure~\ref{fig:benchmark_1}. However, these simplified proxy measures do not generalize well to complex setups, such as comparing prototype representations of two different models or their ensembles~\citep{Keswani_2022_Proto2ProtoCanyou}, as well as comparing with competing approaches.
Overall, there is no consensus in the community on how to evaluate the quality of explanations derived from \ppms~\citep{Nauta2023_wcxai_co-12-for-prototype-models,Nauta2023_csur_evaluating-xai-survey}. 
Evaluation with domain experts on real target tasks is considered the ``best way'' to assess explanations~\citep{doshi2017towards}, but is rarely used by the community due to its prohibitive cost and effort~\citep{Nauta2023_csur_evaluating-xai-survey}.
Finally, the lack of attribute datasets (with object part annotations) hinders both qualitative and quantitative evaluation of these architectures~\citep{Ruis_2021_IndependentPrototypePropagation}.

\summarybox{colorm}{{\bf Summary of Challenge:} \ppms are not grounded in the theoretical foundations of different modalities.  Their training is complex and unstable due to multiple training steps and additional hyperparameters. Their interpretability is difficult to assess due to the lack of well-established evaluation metrics and realistic benchmark datasets.
}

\subsection{\texorpdfstring{\markg Generalization}{Generalization}}

\label{ssec:chall:gen}

\ppms contain specific architectural solutions that limit their real-world applications.

\subsubsection{\texorpdfstring{\markga Tasks}{Tasks}.}

\label{sssec:chall:gen:tasks}
\ppms have been developed for fine-grained image classification, but their application to broader domains remains limited~\citep{Xue_2024_ProtoPFormerConcentratingPrototypical}.
In particular, they have not been tested on small datasets~\citep{Song_2024_MorphologicalPrototypingUnsupervised} and are typically designed for low-resolution images, limiting their clinical applications\footnote{This limitation primarily stems from backbones pre-trained on ImageNet with 224px$\times$224px, whereas clinical images range from mega- (mammography) to even giga-pixels (whole slide imaging).}~\citep{Carmichael_2024_ThisProbablyLooks}.
In addition, they focus on single-label classification~\citep{Ruis_2021_IndependentPrototypePropagation} and have not been well tested in multi-label and multimodal~\citep{Rymarczyk_2023_ProtoMILMultipleInstance} settings, which often occur in real-world applications. Moreover, there is only preliminary work on applying prototypes to more challenging scenarios, such as continual learning~\citep{Rymarczyk_2023_ICICLEInterpretableClass}, the open-world problem~\citep{Zheng_2024_PrototypicalHash}, partial label learning~\citep{Carmichael_2024_ThisProbablyLooks}
, or zero shot classification~\citep{Ruis_2021_IndependentPrototypePropagation}. Finally, these models are often developed without feedback from domain experts~\citep{Fauvel_2023_LightweightEfficientExplainablebyDesign} and do not take into account contextual information (such as time and historical interactions) that is useful for, e.g., email classification~\citep{Wang_2023_PROMINETPrototypebasedMultiView}.

\subsubsection{\texorpdfstring{\markga Assumptions}{Assumptions}.}
\label{sssec:chall:gen:assum}
The bottleneck nature of \ppms plays a critical role in ensuring interpretability by aligning model representations with hu\-man-under\-stand\-able concepts. However, this assumption can also limit the ability of the model to capture complex data patterns~\citep{Zheng_2024_PrototypicalHash}. As a result, \ppms do not support counting the occurrences of prototypes~\citep{Nauta_2023_PIPNetPatchBasedIntuitive}(see Figure~\ref{fig:ppm-reasoning},C), do not capture relationships between detected prototypes~\citep{Zhang_2023_Learningselectprototypical}, and are unable to represent prototypes as hierarchical structures~\citep{Wang_2021_InterpretableImageRecognition}.
In addition, \ppms rely on a fixed number of prototypes~\citep{Barnett_2021_casebasedinterpretabledeep}, and there are no mechanisms to guide the model toward user-desired concepts without risking data leakage~\citep{Bontempelli_2023_ConceptlevelDebuggingPartPrototype}.

\summarybox{colorg}{{\bf Summary of Challenge:} The implicit inductive bias of \ppms limits their ability to learn complex patterns and relationships between prototypes. Their application is currently limited to single-modality, single-label classification tasks with large training datasets, and, in the case of vision \ppms, to images with a low resolution. }

\begin{figure*}[th]
 \centering
        \includegraphics[trim={1cm 1cm 1cm 0cm}, clip, width=\textwidth]{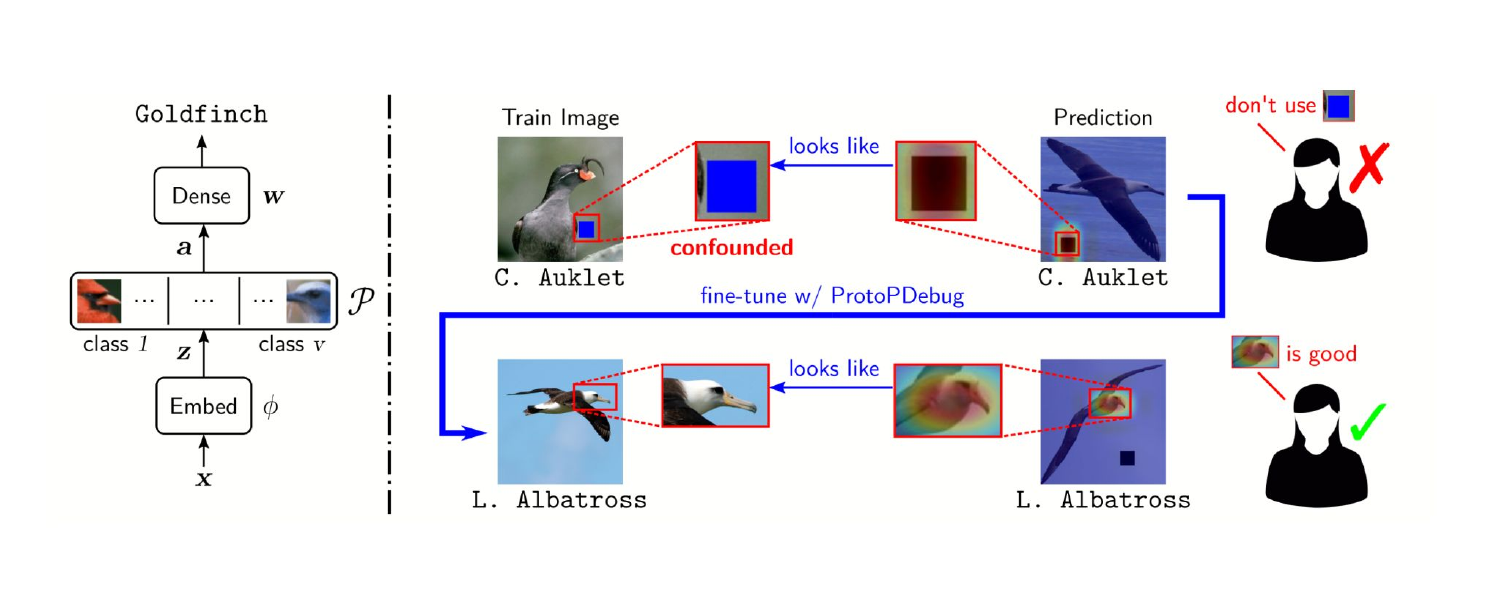}
        \caption{The \textsc{ProtoPDebug} framework~\citep{Bontempelli_2023_ConceptlevelDebuggingPartPrototype} adds human-in-the-loop debugging to \ppms: users inspect explanations, flag misleading or confounded prototypes, and provide feedback to suppress them. The model updates its representations to improve generalization and better align with user-defined concepts. However, human feedback can reduce performance when learned features diverge from human intuition~\citep{Li_2024_ImprovingPrototypicalVisual} or when supervision is misleading or adversarial~\citep{Bontempelli_2023_ConceptlevelDebuggingPartPrototype}.}
    \label{fig:safety_1}
\end{figure*}

\begin{figure*}[th]
 \centering
        \includegraphics[width=\textwidth]{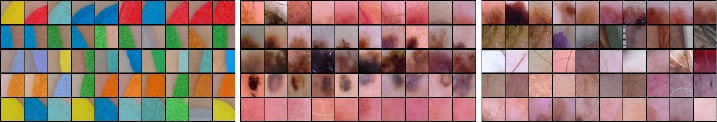}
        \caption{Example of shortcut learning for classifying skin lesions as benign or malignant. Each row shows one prototype: benign (left, center) and malignant (right). \textsc{PiP-Net}~\citep{Nauta_2024_InterpretingCorrectingMedical} learns class-relevant prototypes (center, right) but also shortcut prototypes based on colored patches unique to benign lesions (left).}
    \label{fig:shortcut_learning}
\end{figure*}

\subsection{\texorpdfstring{\markf Safety and Use in Practice}{Safety and Use in Practice}}
\label{ssec:chall:safety}

Human-model interaction has been identified as an open challenge of \ppms~\citep{Nauta2023_wcxai_co-12-for-prototype-models} and has been applied to improve the semantics of prototypes~\citep{Bontempelli_2023_ConceptlevelDebuggingPartPrototype,Li_2024_ImprovingPrototypicalVisual}.
However, while interaction provides valuable human feedback, it carries the risk of reducing the predictive accuracy of the model if the learned features of the model do not match human intuition~\citep{Li_2024_ImprovingPrototypicalVisual}, or of corrupting the model if the human supervision provided is adversarial~\citep{Bontempelli_2023_ConceptlevelDebuggingPartPrototype} (see Figure~\ref{fig:safety_1}).

Other safety concerns are training data bias, to which \ppms are vulnerable~\citep{Carmichael_2024_ThisProbablyLooks}, adversarial attacks that may compromise the model's decision making process~\citep{Rymarczyk_2023_ProtoMILMultipleInstance} and susceptibility to shortcut learning (see Figure~\ref{fig:shortcut_learning}).

In addition to safety concerns, \ppms face several challenges that make them unusable in practice: The learned prototypes currently struggle to generalize to cases outside of the training data~\citep{Wang_2023_PROMINETPrototypebasedMultiView}, and are not always helpful in understanding the model's prediction. Furthermore, models lack intuitive interfaces to visualize the predictions in a user-friendly way~\citep{Wang_2023_PROMINETPrototypebasedMultiView}. 
There is also limited exploration of possible future application areas of \ppms with domain experts~\citep{Fauvel_2023_LightweightEfficientExplainablebyDesign}.   

\summarybox{colorf}{{\bf Summary of Challenge:} To enable safe practical use, there is a need for interactive interfaces that support human inspection and feedback. Moreover, safeguards must be developed to prevent models from adapting to adversarial or misleading human supervision.}

\subsection{Current Progress}
\label{sec:Overview_ideas_tables}

\begin{table}[htb]
\scriptsize
\renewcommand{\arraystretch}{1.5}
\centering
\caption{Ideas for open challenges in \markp Prototypes. }
\label{tab:ideas:prototypes}
\begin{tabular*}{\textwidth}{@{}>{\raggedright\arraybackslash}>{\raggedright\arraybackslash}p{5cm}>{\raggedright\arraybackslash}p{7.2cm}@{}}
\toprule
\textsc{Problem} & \textsc{Key Idea} \\
\midrule
\multicolumn{2}{c}{\texorpdfstring{\colorbox[RGB]{128, 223, 223}{\textsc{Prototypes \textrightarrow{} Number}}}{Number}}\\\addlinespace
Large number of prototypes.
	& Share prototypes among classes. (\textsc{ProtoPShare}~\cite{Rymarczyk_2021_ProtoPSharePrototypicalParts})\\
    & Focal similarity to learn salient prototypes. Share prototypes among classes using soft activations. (\textsc{ProtoPool}~\cite{Rymarczyk_2022_InterpretableImageClassification})\\
	& Share prototypes among classes by using a decision tree instead of a linear model in the final layers. (\textsc{ProtoTree}~\cite{Nauta_2021_NeuralPrototypeTrees})\\
	& Prune rarely used prototypes after training.  (\textsc{ProtoryNet}~\cite{Hong_2023_ProtoryNetInterpretableText})\\
Large number of (redundant/irrelevant) prototypes.
	& Allow shared prototypes among classes. Sparsity loss to minimize the number of prototypes. (\textsc{PiPNet}~\cite{Nauta_2023_PIPNetPatchBasedIntuitive})\\
Prototypes represent redundant concepts.
	& Instead of being a trainable parameter, a prototypes is the average feature vector of clustering feature patches (\cite{Zhu_2025_CVPR})\\
\midrule
\multicolumn{2}{c}{\texorpdfstring{\colorbox[RGB]{191, 239, 239}{\textsc{Prototypes \textrightarrow{} Quality \textrightarrow{} Semantics}}}{Semantics}}\\\addlinespace
Prototypes encode background information.
	& Increase the number of classes for \ppm training by creating pseudoclasses to increase model capacity and generate more meaningful prototypes. (\textsc{PCPPN}~\cite{Choukali_2024_Pseudoclasspartprototype})\\
	& Shapley value computation as replacement for prototype similarity in the prototype layer to satisfy theoretical properties. (\textsc{ProtoPFaith}~\cite{Wolf_2024_KeepFaithFaithful})\\
Visualisations of prototypes are ambiguous when the semantics of the prototype is unclear.
	& Multiple visualizations for each prototype. (\textsc{ProtoConcepts}~\cite{Ma_2023_ThisLooksThose})\\
Prototypes are not meaningful and do not align with human preferences.
	& Reward model based on human feedback for prototype selection. (\textsc{R3 framework}~\cite{Li_2024_ImprovingPrototypicalVisual})\\
Prototypes lack semantics because they do not explicitly account for geometric transformations or pose variations.
	& Deformable convolutional layers generate offsets that are used by the prototypes to adjust their spatial location. (\textsc{Deformable ProtoPNet}~\cite{Donnelly_2022_DeformableProtoPNetInterpretable}) \\
Prototypes are ambiguous and represent  entangled concepts.
	& Represent prototypes in a  feature space spanned by basis concepts. (\textsc{TesNet}~\cite{Wang_2021_InterpretableImageRecognition})\\
Prototypes are not meaningful or are irrelevant for humans.
	& Learn invertible representations of prototypical distributions. (\textsc{ProtoFlow}~\cite{Carmichael_2024_ThisProbablyLooks})\\
\ppms learn prototypes representing shortcuts (confounders). 
	& An interactive framework incorporates user feedback into  the model. Prevent re-learning prototypes that represent confounders by an additional loss term. (\textsc{ProtoPDebug}~\cite{Bontempelli_2023_ConceptlevelDebuggingPartPrototype})\\
Prototypes have fixed shape and size and can therefore be ambiguous.
	& Deform prototypes into (smaller) sub-prototypes and filter out irrelevant sub-prototypes. (\textsc{ProtoViT}~\cite{Ma_2024_Interpretableimage})\\
Prototypes have fixed size and shape and thus do not represent (medical) concepts well.
	& Include representation of spatial location of prototypes (occurrence) map and allow prototypes of dynamic size. (\textsc{XProtoNet}~\cite{Kim_2021_XProtoNetDiagnosisChest})\\
\midrule
\multicolumn{2}{c}{\texorpdfstring{\colorbox[RGB]{191, 239, 239}{\textsc{Prototypes \textrightarrow{} Quality \textrightarrow{} Similarity}}}{Similarity}}\\\addlinespace
Prototype activations are not aligned with locations in input feature space.
	& Replace upsampling with Shapley value attributions of pixels to prototypes and visualize pixel heatmaps. (\textsc{ProtoPFaith}~\cite{Wolf_2024_KeepFaithFaithful})\\
Prototypes are ambiguous.
	& Contrastive loss term in the loss function and pretrain the feature extraction backbone in a self-supervised manner. (\textsc{PiPNet}~\cite{Nauta_2023_PIPNetPatchBasedIntuitive})\\
	& Spatially align deep feature maps with image locations. 
    Enforce class-specific prototypes. (\cite{Huang_2023_EvaluationImprovementInterpretability})\\
    & Disentangle color from shape and texture in prototype learning.(\textsc{LucidPPN}~\cite{Pach_2024_LucidPPNUnambiguousPrototypical})\\
Prototype similarity in latent space is not reflecting similarity in image space.
	& Learn invertible representations of prototypical distributions. (\textsc{ProtoFlow}~\cite{Carmichael_2024_ThisProbablyLooks})\\
Prototype visualizations in test images are misleading.
	& Alignment loss enforces similarity of the visualization in the original image to that of an image with only the prototypical part unmasked. (\cite{Sacha_2024_Interpretabilitybenchmarkevaluating})\\
\bottomrule
\end{tabular*}
\end{table}

\begin{table}[htb]
\scriptsize
\renewcommand{\arraystretch}{1.5}
\caption{Ideas for open challenges in \markm Methodology and \markg Generalization}.
\label{tab:ideas:method-general}
\begin{tabular*}{\textwidth}{@{}>{\raggedright\arraybackslash}>{\raggedright\arraybackslash}p{5cm}>{\raggedright\arraybackslash}p{7.2cm}@{}}
\toprule
\textsc{Problem} & \textsc{Key Idea} \\
\midrule
\multicolumn{2}{c}{\texorpdfstring{\colorbox[RGB]{156,198,246}{\textsc{Methodology \textrightarrow{} Performance}}}{Performance}}\\\addlinespace
\ppms can not detect OOD data. 
	& Positive linear decision layer and a special normalization term for logits. (\textsc{PiPNet}~\cite{Nauta_2023_PIPNetPatchBasedIntuitive})\\
\ppms underperform on OOD data. 
	& An interactive framework incorporates user feedback into the model. Prevent re-learning prototypes that represent confounders by an additional loss term. (\textsc{ProtoPDebug}~\cite{Bontempelli_2023_ConceptlevelDebuggingPartPrototype}) \\
Predictive performance of \ppms is low compared to black-box models. 
	& Learn support prototypes close to decision boundary and ensemble two \ppms, one with support prototypes and the other with trivial prototypes. (\textsc{ST-ProtoPNet}~\cite{Wang_2023_LearningSupportTrivial})\\
Predictive performance of \ppms is low compared to black-box models.
	& ViT instead of a CNN as backbone and a neural tree decoder for classification. (\textsc{ViT-NeT}~\cite{Kim_2022_ViTNeTInterpretableVision})\\
Predictive performance is low.
	& Use top-k average pooling instead of max-pooling to compute prototype similarity scores. (\textsc{IAIA-BL}~\cite{Barnett_2021_casebasedinterpretabledeep})\\
Unstable predictions. 
	& Data-dependent similarity metric in representation space to have prototypes with similar semantics close in representation space. (\textsc{ProtoPShare}~\cite{Rymarczyk_2021_ProtoPSharePrototypicalParts})\\
\midrule
\multicolumn{2}{c}{\texorpdfstring{\colorbox[RGB]{156,198,246}{\textsc{Methodology \textrightarrow{} Benchmark and Evaluation}}}{Benchmark and Evaluation}}\\\addlinespace
\ppm interpretability is human-centric and not well evaluated. 
	& Introduce human-centric metrics and experiments. (\cite{Davoodi_2023_interpretabilitypartprototypebased})\\
Ambiguous prototypes remain undetected. 
	& Use annotated data to evaluate prototype purity.  (\textsc{PiPNet}~\cite{Nauta_2023_PIPNetPatchBasedIntuitive})\\
Lack of evaluation metrics that quantitatively evaluate the interpretability of prototypes.
	& New interpretability metrics to evaluate consistency and stability of prototypes.(\cite{Huang_2023_EvaluationImprovementInterpretability})\\
Lack of systematic evaluation of human interpretability.
	& Systematic study design and task setup for human-centered evaluation. (\textsc{HIVE}~\cite{Kim_2022_HIVEEvaluatingHuman})\\
Lack of systematic evaluation framework (for radiology use case).
	& Evaluation framework and domain-specific evaluation metrics. (\textsc{PEF-Coh}~\cite{Pathak_2024_PrototypebasedInterpretableBreast})\\
Lack of benchmarks that assess prototype visualizations.
	& Introduce metrics for prototype location, activation, ranking, and accuracy. (\cite{Sacha_2024_Interpretabilitybenchmarkevaluating})\\
\midrule
\multicolumn{2}{c}{\texorpdfstring{\colorbox[RGB]{255,199,166}{\textsc{Generalization \textrightarrow{} Tasks}}}{Tasks}}\\\addlinespace
\ppms are not designed for tasks beyond fine grained classification (e.g., mammograms).
    & Extend the ProtoPNet architecture through the use of fine-grained annotations, modified modular training and multi-stage reasoning. (\textsc{IAIA-BL}~\cite{Barnett_2021_casebasedinterpretabledeep})\\
    & A prototype framework for visual question answering tasks via the learning of question aware visual prototypes.(\textsc{ProtoVQA}~\cite{Diao_2025_ProtoVQAAdaptablePrototypical})\\
\midrule
\multicolumn{2}{c}{\texorpdfstring{\colorbox[RGB]{255,199,166}{\textsc{Generalization \textrightarrow{} Assumptions}}}{Assumptions}}\\\addlinespace
Prototypes for sequential data lack interpretability.
    & Neural network architecture incorporating a prototype selection strategy for sequential data. (\textsc{SESM}~\cite{Zhang_2023_Learningselectprototypical})\\
\bottomrule
\end{tabular*}
\end{table}

Several papers from our surveys corpus have addressed some of the open challenges \ppms are facing. To assist in future research, we systematized the ideas and present them concisely. We provide an overview of ideas addressing \markp \textbf{Prototypes} in Table~\ref{tab:ideas:prototypes}. Table~\ref{tab:ideas:method-general} shows an overview of recent advances in terms of  \markm \textbf{Methodology} and \markg \textbf{Generalization}. Work that addresses \markf~\textbf{Safety and Use in Practice} is not present in our corpus.

\section{Research Directions}
\label{sec:directions}

\begin{figure*}[tbp]
 \centering        \includegraphics[width=1.0\linewidth]{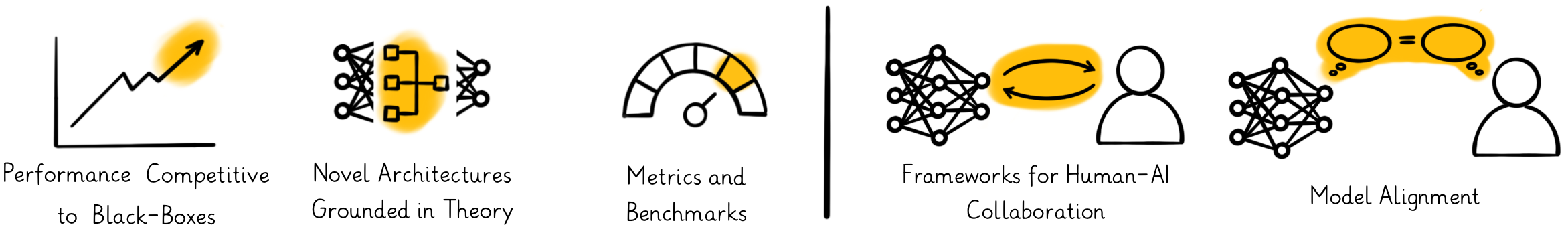}            \caption{Principle directions for future research. In addition to technical and theoretical research (left), it is important to address human-centered issues (right).}
    \label{fig:directions}
\end{figure*}

With progress in parts of selected challenges (cf. Tables \ref{tab:ideas:prototypes} and \ref{tab:ideas:method-general}), 
we synthesized five main research directions for future work (see Figure~\ref{fig:directions}) on part-prototype models (\ppms).
We describe each direction, outline some ideas and note which challenges they address.

\subsection{Performance Competitive to Black Boxes}
\label{ssec:directions:performance}
\ppms are limited in their expressiveness because they make decisions based on a fixed number of fixed-size localized image features (prototypes) and do not model their interrelationships. To increase the expressiveness of the model, future work should focus on relaxing these constraints. For example, multiple layers of the backbone CNN can be used to obtain prototypes corresponding to different types of visual features, such as color, shape, and higher-level object parts~\citep{Wang_2024_MCPNetInterpretableClassifier}. The prototypes should also be of different sizes and shapes as shown in previous work~\citep{Donnelly_2022_DeformableProtoPNetInterpretable}. 
Other research directions could focus on modeling spatial relationships between prototypes, e.g., using graph neural networks, and modeling hierarchical relationships to recognize objects even when they are partially occluded. Finally, the number of prototypes can be automatically optimized during training by adding and removing prototypes using ideas from pruning techniques.

Research in these directions addresses the challenges \markma~\textbf{Performance} 
and \markpa \textbf{Number of prototypes}, %
and would make \ppms more likely to be adopted by stakeholders who prioritize performance over interpretability in ML systems.

\subsection{Novel Architectures Grounded in Theory}
\label{ssec:directions:architectures}
\ppms are not intrinsically interpretable in practice, because prototypes are learned in an unsupervised manner, and therefore lack human-understandable semantics.
An important direction for future research is to clearly define prototypes for all modalities. The most concrete definition is available for vision tasks: prototypes are commonly defined as ``visual concepts'' corresponding to any part of the image to which humans could assign a label (e.g., feather). However, there is no clear definition of a visual concept with respect to the optimal level of granularity (e.g., a barb is part of a feather is part of a wing is part of a bird). Generally, the definition is that a prototypical concept is an element that is ``a reasonably small and sufficiently large part of the input that has some meaning to humans'' and is hardly actionable.
Rigorous definitions of prototypes can be derived from findings in human information processing, such as human visual perception, linguistics~\citep{panther2008prototype}, and cognitive science. Based on these definitions, we could then design novel \ppm architectures that reflect the human part-of relation, and guide the model (e.g., with few annotations) to learn the correct type of prototypes for different modalities.
\ppms for modalities other than vision are currently rare (cf. table at the top of Figure~\ref{fig:chart}). Hence, more guidelines on what constitutes a prototype for a certain modality could foster the development of not only single-modal models for other modalities but also multi-modal models. One of such models, recently proposed by \cite{DeSanti2024_xailbr_pip3d-alzheimer}, successfully combines prototypes for structured patient data (age) and 3D imaging data (brain computer tomography) into a multi-modal \ppm.

A clearer definition of prototypes would not only address the current criticism that prototypes are not interpretable (\markpb~\textbf{Semantics})
and ground models in theory (\markma~\textbf{Theoretical Foundation})
but also potentially improve  \markma~\textbf{Performance},
and make models more trustworthy and applicable in a wider range of scenarios (\markga~\textbf{Tasks},
\markf~\textbf{Safety and Use in Practice}).

\subsection{Frameworks for Human-AI Collaboration}
\label{ssec:directions:interactive}
\ppms are not always useful in practice, either due to common ML problems (e.g., OOD generalization, shortcut learning~\citep{Geirhos2020_ShortcutLearningDeep}) or due to learning features that are counterintuitive or irrelevant for humans. 
An important direction is to extend the work on interactive \ppms (prior work by \cite{Bontempelli_2023_ConceptlevelDebuggingPartPrototype,Li_2024_ImprovingPrototypicalVisual}) to allow domain experts to adjust the reasoning process of the models. This includes adding relevant prototypes, removing irrelevant or redundant prototypes, removing prototypes corresponding to shortcuts, and adjusting the weights of prototypes.
It would also be interesting to inject domain knowledge prior to training to provide the model with a priori guidance on what is (or is not) important for the task from a human perspective.

We believe that adaptable \ppms would increase user trust (\markf~\textbf{Safety and Use in Practice})
since they would align more with human reasoning (\markpb~\textbf{Semantics}, %
\markga~\textbf{Assumptions}),
help keep models up-to-date, improve out-of-distribution generalization, and potentially improve overall model \markma~\textbf{Performance}.

\subsection{Model Alignment}
\label{ssec:directions:alignment}

For humans, some prototypes and activations do not look similar (see Section~\ref{ssssec:chall:proto:quality:sim}, and Figure~\ref{fig:overview}). The similarity of prototypes is computed in latent space, which does not guarantee fidelity~\citep{Xu-Darme_2023_Sanitycheckspatch}, and it remains unclear what the similarity score represents: Is the color important or the shape? Or maybe they both should be ignored? Furthermore, inaccuracies in the visualization of prototypes hamper interpretability~\citep{Carmichael_2024_PixelGroundedPrototypicalPart, Gautam_2023_ThislooksMore}.
Future research should focus on metrics beyond cosine similarity, which is an unreliable metric for similarity~\citep{Steck_2024_CosineSimilarityEmbeddingsReally}. 
To explain similarities of an input to a learned prototype, natural language explanations (e.g., ``this is a bird's head with orange and blue coloring'') can be generated using vision-language models~\citep{feldhus_2023_SaliencyMaps}.
However, those are post-hoc rationals, not necessarily faithful to the model's internal decision process, and only show what the vision-language model ``sees'' in a prototype.
An alternative option that is faithful to the model is to use controlled perturbations in the input space (e.g., changing the color) and assess whether the prototype would still be activated~\citep{Nauta_2021_ThisLooksThat}, or derive visualizations that capture the essence of the prototypes, such as sketching, background removal, or improved localization). Furthermore, disentangling the different visual features (color, shape, and texture) and learning prototypes that represent each information separately could improve the clarity of explanations.
Finally, an interesting direction would be to combine the interpretability of semantic features as used in concept-bottleneck models~\citep{Koh2000_concept-bottleneck-models} and the ``part-of'' idea of \ppms as exemplified by \cite{Wan2024_interpretable-object-recognition-semantic-prototypes} with semantically annotated prototypes.

Research in this direction aims to improve the model's alignment with humans (\markpb~\textbf{Similarity}) %
and to increase trust for using them safely in applications (\markf~\textbf{Safety and Use in Practice}).
Developing human-aligned similarity metrics would have implications beyond \ppms.

\subsection{Metrics and Benchmarks}
\label{ssec:directions:benchmarks}
\ppms claim to be human understandable. However, human understanding depends on the expertise and background knowledge of specific stakeholders and is inherently difficult to measure~\citep{Boogert_2018_Measuringunderstandingindividual}. This is especially true given that XAI, as a relatively young field of research, does not have an established evaluation methodology~\citep{Nauta2023_csur_evaluating-xai-survey}.
Therefore, an important next step is to consolidate evaluation metrics. This comprises the comparison, revision and adaption of existing metrics including standard XAI evaluation metrics~\citep{Nauta2023_csur_evaluating-xai-survey}, specific metrics for \ppms~\citep{Nauta2023_wcxai_co-12-for-prototype-models,Huang_2023_EvaluationImprovementInterpretability}, and domain-specific metrics (e.g., ~\citep{Pathak_2024_PrototypebasedInterpretableBreast}).
Specifically designed synthetic benchmarks, such as the FunnyBirds dataset for fine-grained image recognition~\citep{Hesse_2023_FunnyBirdsSyntheticVision}, and additional benchmark datasets for other machine learning tasks and input modalities could provide insight into \ppms' failure modes and support model improvement. 
For faster research cycles, metrics and datasets should be integrated into well-established evaluation framework~\citep{Le2023_ijcai_benchmarking-xai}.

Research in this direction directly addresses the challenge of evaluating \ppms (\markma~\textbf{Benchmark and Evaluation}),
and contributes to validating and improving (\markma~\textbf{Performance}). %
Moreover, domain-specific and problem-centric metrics and benchmarks would make \ppms more trustworthy and safer to use in applications (\markf~\textbf{Safety and Use in Practice}).

\section{Conclusion}
\label{sec:conclusion}

Since their inception in 2019, with an initial application in fine-grained image recognition, part-prototype models (\ppms) have seen the development of multiple extensions and variations in modalities beyond vision (text, sequence, graph, sound, video). 
They have been applied in several application domains, particularly in those where interpretability is valued (e.g., medicine, finance).

Despite being intrinsically interpretable, our analysis shows \ppms still suffer from multiple challenges (e.g., low quality of prototypes, lack of theoretical foundation, and non-competitive predictive performance), making them less likely to be used than black box models. 

For future work, we provide five research directions and outline concrete research ideas.
This includes the development of interactive frameworks for human-AI collaboration to address the semantic shortcomings of prototypes, and the design of novel theory-based architectures to address the lack of theoretical foundation in \ppms. 
In addition, aligning models with human reasoning by introducing human-aligned similarity metrics and disentangling the different visual features (color, shape and texture) would improve their usefulness in practice. 

We envision this survey as a useful resource for researchers who are interested in alternatives to black box models, and we hope that the research directions we provide will pave the way for better \ppms, ultimately providing different ML stakeholders with accurate and interpretable models.

\begin{credits}
\subsubsection*{\ackname} This research was partially funded by the Deutsche Forschungsgemeinschaft (DFG, German Research Foundation) – project number 536124560. Work of Adam Wróbel and Bartosz Zieliński was founded by ``Interpretable and Interactive Multimodal Retrieval in Drug Discover'' project. The ``Interpretable and Interactive Multimodal Retrieval in Drug Discovery'' project (FENG.02.02-IP.05-0040/23) is carried out within the First Team programme of the Foundation for Polish Science co-financed by the European Union under the European Funds for Smart Economy 2021-2027 (FENG).
Work of Tomasz Michalski was funded by National Science Centre (Poland) grant number 2022/47/B/ST6/03397.
\end{credits}

\bibliographystyle{splncs04}
\bibliography{bibliography_short.bib}
\end{document}